\documentclass{IOS-Book-Article}

\usepackage{mathptmx}
\usepackage{soul}\setuldepth{article}
\usepackage{multirow}
\usepackage[table]{xcolor}
\usepackage{booktabs}

\usepackage{tabularx}
\usepackage{makecell}
\def\hb{\hbox to 11.5 cm{}}
\usepackage[skip=2pt,font=scriptsize]{caption}

\usepackage{microtype}
\usepackage{graphicx}
\usepackage{amsmath}
\usepackage{lipsum}
\usepackage[title]{appendix}

\DeclareCaptionLabelSeparator{custom}{ }
\DeclareCaptionFormat{custom}
{%
    \textbf{#1#2}{\small #3}
}
\usepackage[nohyperlinks,nolist]{acronym}
\begin{acronym}[XXX]
    \acro{LLM}{Large Language Model}
    \acro{NLP}{Natural Language Processing}
    \acro{DPR}{Dense Passage Retrieval}
\end{acronym}

\begin{document}

\pagestyle{headings}
\def\thepage{}
\begin{frontmatter}              

\title{Reframing Tax Law Entailment as Analogical Reasoning}

\markboth{}{October 2023\hb}

\author[A]{\fnms{Xinrui} \snm{Zou}$^*$\orcid{0009-0002-0446-493X}
            \thanks{Corresponding Author: Xinrui Zou, xzou8@jhu.edu}
},
\author[A]{\fnms{Ming} \snm{Zhang}$^*$\orcid{0009-0007-5079-9382}},
\author[A]{\fnms{Nathaniel} \snm{Weir}\orcid{0000-0002-6734-3214}},
\author[A]{\fnms{Benjamin} \snm{Van Durme}\orcid{0000-0003-4328-4288}}
and
\author[B]{\fnms{Nils} \snm{Holzenberger}\orcid{0000-0002-0844-1391}}

\runningauthor{X. Zou \& M. Zhang et al.}
\address[A]{Johns Hopkins University}
\address[B]{T\'el\'ecom Paris, Institut Polytechnique de Paris}

\let\svthefootnote\thefootnote
\let\thefootnote\relax\footnotetext{$^*$These two authors contributed equally to this work.}

\begin{abstract}

Statutory reasoning refers to the application of legislative provisions to a series of case facts described in natural language.
We re-frame statutory reasoning as an analogy task, where each instance of the analogy task involves a combination of two instances of statutory reasoning.
This increases the dataset size by two orders of magnitude, and introduces an element of interpretability.
We show that this task is roughly as difficult to \acl{NLP} models as the original task.
Finally, we come back to statutory reasoning, solving it with a combination of a retrieval mechanism and analogy models, and showing some progress on prior comparable work.

\end{abstract}

\end{frontmatter}
\markboth{October 2023\hb}{October 2023\hb}

\section{Introduction}
\label{introduction}

In the legal domain, predicting case outcomes is a critical task that requires multiple complex skills.
One of them is statutory reasoning, the ability to understand statutes and to apply them to the facts of a case.
Prior work has introduced the StAtutory Reasoning Assessment dataset (SARA), framing statutory reasoning as a textual entailment problem~\cite{Holzenberger2020ADF}.
Here, we re-frame statutory reasoning as an analogy problem.

\begin{figure}
    \centering
    \includegraphics[width=\textwidth]{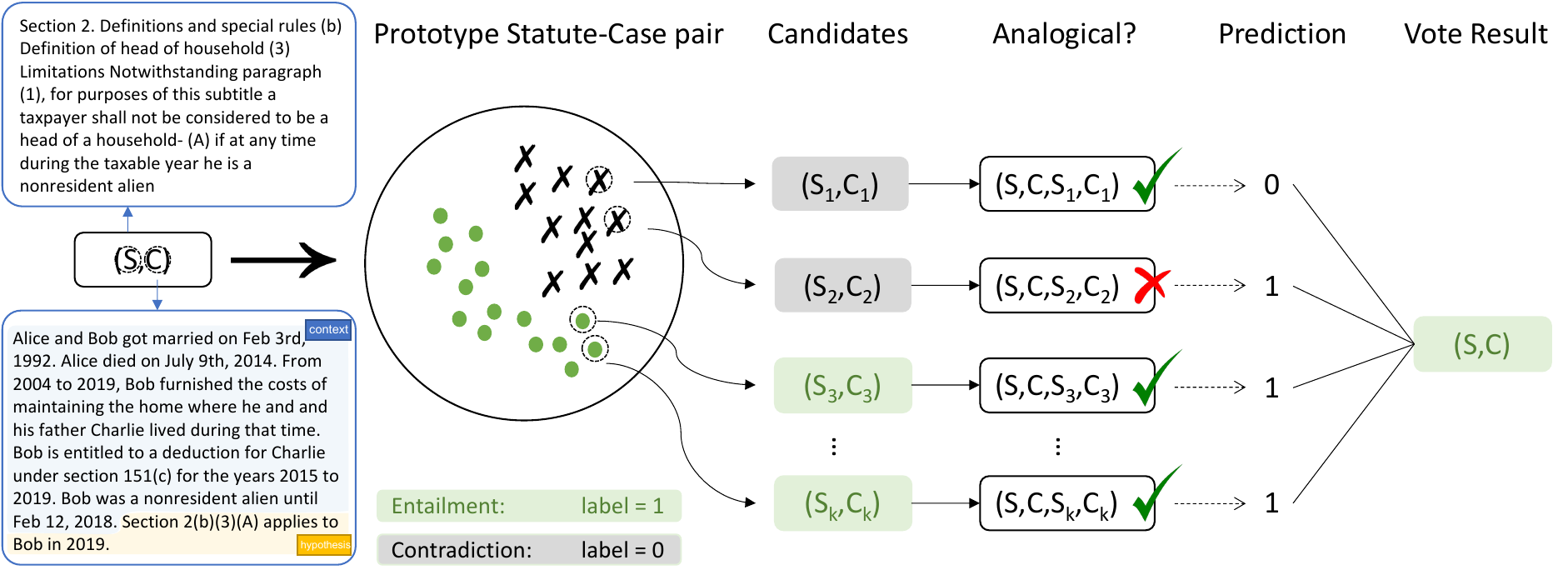}
    \caption{
    Performing statutory reasoning as a combination of retrieval and analogy.
    To determine the label of $(S,C)$, candidate statute-case pairs are retrieved from a set of prototypes, and used to form quadruples.
    The quadruples are labeled as analogical or not.
    Analogy labels are aggregated through voting to yield labels for statutory reasoning.
    }
    \label{fig:a2e}
\end{figure}

In the SARA dataset, each data sample consists of a case and a relevant statute, where a case is a set of facts described in natural language, and statutes refer to the relevant sections adapted from US Federal Tax Law (as seen in Figure~\ref{fig:a2e}, leftmost example).
Each example is either labeled \emph{entailment} or \emph{contradiction}, depending on whether the statute applies to the case.

The SARA dataset considers cases in isolation.
In contrast, most legal systems emphasize the interdependency of cases.
For example, consistency of decisions across cases is a desirable property of a legal system.
This form of consistency can be framed as an analogical relationship between two pairs~\cite{Lim2019SolvingWA}.
This relationship is expressed as \emph{a:b::c:d}, representing \emph{a is to b as c is to d};
for example, \emph{the calf is to the cow as the foal is to the mare}.

We extend this concept to the domain of statutory reasoning, wherein $S$ and $C$ denote statutes and cases, respectively.
Instead of focusing solely on the relationship between $S$ and $C$ within one data sample, we put two pairs together to form a \emph{quadruple} $(S_1, C_1, S_2, C_2)$, so as to capture the relationship between the two statute+case pairs $(S_1, C_1)$ and $(S_2, C_2)$.
We label the quadruple as \emph{analogy} if both pairs in $S_1:C_1::S_2:C_2$ share the same entailment label (\emph{entailment}/\emph{contradiction}), and \emph{not analogy} otherwise.

The rationale behind forming these analogies is that different statutes and cases might have inherent correspondences that an NLP model can detect and leverage. For instance, tax rates mentioned in one statute (e.g., §1) might have similarities to those in another, differing mainly in value. Subparagraphs in statutes like §152(d)(2) and their affiliated cases hint at a pattern of similarity, suggesting that analogies based on word swaps could be potent. Moreover, for the analogy to be meaningful and robust, it's crucial that the two statutes being compared, $S_1$ and $S_2$, are not identical. As indicated by \cite{Lim2019SolvingWA}, when $S_1$ equals $S_2$, the analogy breaks down. Taking an example from word analogies, “king is to king as man is to woman” is inherently misleading. More formally, the case S1=S2 can be represented as S1: C1 :: S1 : C2, leading to C1:C2 :: S1: S1, implying C1 = C2.

By employing this methodology, we can increase the size of the dataset from $N$ to $\binom{N}{2}$.
We evaluate multiple methods on this task, utilizing accuracy as the metric, defined by the proportion of instances where the methods accurately predict the \emph{analogy}/\emph{not analogy} label.
While the SARA dataset also contains a set of cases requiring numerical answers, we focus on the entailment component of the SARA dataset, as it lends itself best to being reframed as an analogy problem.

\begin{figure}
    \centering
    \includegraphics[width=0.5\textwidth]{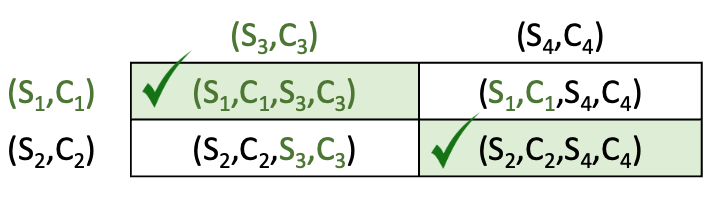}
    \caption{
    Data generation.
    $(S_1, C_1)$ and $(S_3, C_3)$ are pairs from the SARA dataset labeled \emph{entailment};
    $(S_2, C_2)$ and $(S_4, C_4)$ are labeled \emph{contradiction}.
    When forming quadruples, $(S_1, C_1, S_4, C_4)$ and $(S_2, C_2, S_3, C_3)$ are labeled \emph{not analogy} (white cells);
    $(S_1, C_1, S_3, C_3)$ and $(S_2, C_2, S_4, C_4)$ are labeled \emph{analogy} (green cells).
    }
    \label{fig:fig2}
\end{figure}

Furthermore, we use trained analogy models to solve the original textual entailment problem.
In doing so, we simulate the process of looking at similar prior cases for deciding a new case.
Given a new example $(S_t, C_t)$, we retrieve top similar pairs $(S_r, C_r)$ from our prior cases with known entailment labels to constitute new analogy tuples, $(S_t, C_t, S_r, C_r)$.
We predict the entailment label of $(S_t, C_t)$ based on analogy labels predicted for $(S_t, C_t, S_r, C_r)$, as illustrated in Figure~\ref{fig:fig2}.
The use of analogies and the retrieval of similar cases independently provide a form of interpretability to the model's decision, that is absent from the direct application of \acp{LLM} and from most statistical models.

\section{Related Work}

\subsection{ Statutory Reasoning }

Prior approaches to statutory reasoning range from expert-constructed symbolic systems to large neural language models.
Expert systems for legal tasks rely on the manual translation of rules into code, to build knowledge-based systems~\cite{McCarty1977ReflectionsO}.
Examples include the formalization of part of the Canada Income Tax Act~\cite{Sherman:87} and of the 1987 British Nationality Act~\cite{sergot86british}, as well as the SARA dataset.
By construction, these systems are highly accurate, consistent, and auditable.
However, they rely heavily on the accumulated experience of experts and they do not have any self-learning capabilities, making their extension to new legislation and cases labor-intensive.

At the other end of the spectrum, machine reading models based on statistical methods are trained with labeled data, with the goal of generalizing to unseen cases and legislation.
Direct application of BERT~\cite{devlin19bert} to SARA only garners marginal improvement over a majority baseline~\cite{Holzenberger2020ADF}, mostly due to scarce training data.
Subsequent studies have further improved upon this baseline from two distinct perspectives.
First, the integration of GPT-3~\cite{brown20language}, jointly with specific prompting techniques such as Chain-of-Thought~\cite{wei2022chain}, and zero-shot reasoning~\cite{kojima2022large}, resulted in a significant improvement over prior work~\cite{blair2023can}.
Numerous \acp{LLM} have been evaluated on the SARA dataset, with varying success~\cite{guha23legalbench}, suggesting that performance depends on model size and possibly on pre-training data.

Second, with the SARA v2 dataset, statutory reasoning is decomposed into four sequential tasks~\cite{Holzenberger2021FactoringSR}, in order to reduce the complexity associated with tackling statutory reasoning in a single reasoning step.
In the same vein, some methods consider the law as a series of provisions and use \ac{NLP} techniques to semantically classify and extract rules from legal documents~\cite{Biagioli2005AutomaticSE,Dragoni2016CombiningNA}.
The aim is to extract semi-structured information from legal language using shallow parsing.
These approaches are mainly helpful for legal analysts and do not achieve automatic statutory reasoning in the stricter sense.

\subsection{ Analogical Reasoning }

Analogical reasoning has been used in a number of AI tasks~\cite{prade21analogical}: machine translation~\cite{lepage05purest}, computer vision~\cite{law17learning}, solving visual puzzles~\cite{ichien21visual}, learning to rank~\cite{fahandar21analogical}, \emph{inter alia}.
In \ac{NLP}, analogical reasoning is often framed as a geometrical relationship between high-dimensional vector embeddings such as Word2vec~\cite{mikolov2013efficient} and GloVe~\cite{pennington2014glove}.
Later on, these methods were extended with the vector offset method, designed to solve word analogy problems:
a modified form of cosine distance between embeddings captures analogical relationships between the words they represent~\cite{mikolov2013linguistic}.
In~\cite{Lim2019SolvingWA}, this method is generalized by turning analogical relationships into a binary classification task: a quadruple of words either is or is not a valid analogy.
In contrast with the geometric view of analogy described earlier, a neural-network-based binary classifier is learned directly from labeled examples. 

Going from words to sentences, analogical reasoning can be used for question answering~\cite{Diallo2019LearningAS} and sentence-level analogies~\cite{mao23embedding},
demonstrating the effectiveness of analogy-based reasoning at the sentence level.
Our work relies on analogical reasoning at the sentence and paragraph level.

In a legal context, statutory reasoning is distinct from case-based reasoning.
Our goal is to reason with statutes, and we try to leverage (dis)similarities between statutes and cases, in contrast to predicting case outcomes based on other cases.
Case-based reasoning is occasionally associated with a form of analogical reasoning.
For example, cases can be represented as a set of high-level features~\cite{bruninghaus03predicting}, so that predicting outcomes involves a comparison of the current case to a set of related cases.
Beyond that, there is a rich literature on case-based reasoning with some form of analogical reasoning~\cite{mumford23combining}.
\section{Methodology}
\label{sec:methodology}

This section presents an analysis of the analogical task for statutory reasoning, which poses a challenge due to the absence of a dedicated dataset.
To overcome this limitation, we generate quadruples in Section~\ref{sec:quadruple-generation}.
Subsequently, the analogical task is evaluated using GPT-3.5~\cite{ouyang2022training} and GPT-4~\cite{OpenAI2023GPT4TR}, which have demonstrated exceptional performance across various language tasks, including statutory reasoning~\cite{blair2023can}.
The vector offset method described above serves as the baseline for this analogy task.
Then, the analogy task is framed as a textual entailment task, so that textual entailment models can be tested.
We also explore solving statutory reasoning by transforming it into an analogy problem, and applying our best-performing analogy models augmented with retrieval mechanisms.

\subsection{Quadruple Generation}
\label{sec:quadruple-generation}

We use the SARA dataset~\cite{Holzenberger2020ADF} to generate examples of analogies, to be used as training and test data.\footnote{This new dataset can be found at: \url{https://nlp.jhu.edu/law/sara-analogy.tar.gz}}
The dataset, which is based on the U.S. Tax code, includes one entailment and one contradiction case for each atomic law in the SARA statutes.
Analogical quadruples can be constructed from these case-statute pairs, as illustrated in Figure~\ref{fig:fig2}.
Positive quadruples comprise two entailment cases, or two contradiction cases.
Negative quadruples comprise one positive and one negative pair.
Within each train, dev and test sets we create analogies by taking all possible combinations of 2 entailment pairs.
By applying this process, the training set expanded from 158 to 12403 examples.\footnote{The dev and test sets expanded from 38 and 80 to 703 and 3160 examples. The original SARA dataset has respectively 158, 18, 100 in train, dev and test set. In order to maintain a reasonable proportion between dev and test set after transforming them into analogy examples, we randomly moved 20 examples from test to dev set.}
Examples of analogical quadruples can be found in Appendix~\ref{sec:quadruple example}.

\subsection{GPT Models on the Analogy Task}
\label{sec:gpt-3.5}
\vspace{-5pt}
We set out to test how the most powerful GPT-3.5~\cite{ouyang2022training} and GPT-4~\cite{OpenAI2023GPT4TR} models currently exposed through public APIs would perform this analogy task, testing a number of prompts. Prompts details can be found in Appendix \ref{sec:gpt3.5_prompts}.

\paragraph{Few-shot.}
For a given test case, we utilize three additional training cases as prompts, each comprising the full text of S1, C1, S2, and C2.
The text of each case in the prompts is separated into two distinct segments, designated as ``Context'' and ``Hypothesis''.
A comparative question~--- ``Question: Is S1 to C1 as S2 to C2?''~--- and the corresponding answer~--- ``A: Yes or No''~--- are appended to these training cases.
Finally, we attach the test case with the same format as the training cases.

\paragraph{Zero-shot.}
We use the test case with the previous format as a prompt without additional training cases.

\paragraph{Chain-of-Thought.}
We use six training examples comprising four sentences labeled as S1, C1, S2, and C2, respectively. A query is attached to each example in the format of ``Is S1 to C1 as S2 to C2?'' and a corresponding chain-of-thought explaining the reasoning for the comparison, concluding with either a positive (``Yes'') or negative (``No'') response.

\subsection{Vector Offset Method}

\setlength{\intextsep}{5pt}%
\setlength{\columnsep}{10pt}%
\begin{figure}
    \centering
    \includegraphics[width=0.55\textwidth]{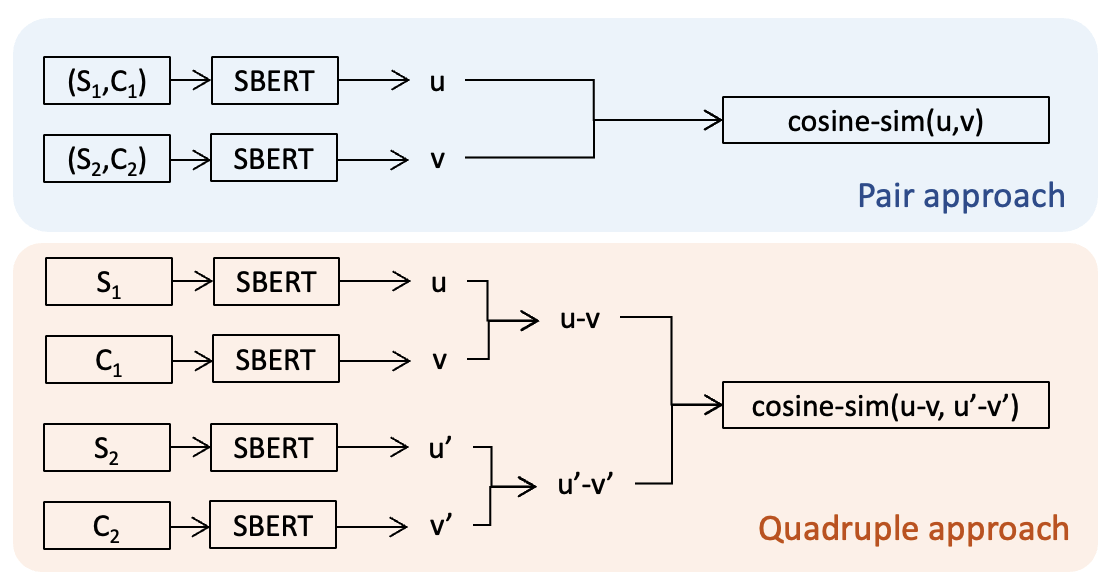}
    \caption{Vector offset methods.}
    \label{fig:similarity_method}
\end{figure}

\newcommand{\fsbert}{f_{\mathrm{sbert}}}

Similar to the modeling assumption of~\cite{mikolov2013linguistic}, the relationship between $(S_1,C_1)$ and $(S_2,C_2)$ can be expressed as a relationship between vectors in embedding space.
We encode each sentence into a fixed-length embedding that represents its semantics, using Sentence-BERT~\cite{Reimers2019SentenceBERTSE}.
Let $\fsbert(x)$ be the embedding of sentence $x$ in vector space that is computed by a Sentence-BERT encoder.
To demonstrate the usefulness of computing pairwise offsets, we empirically test two methods, both illustrated in Figure~\ref{fig:similarity_method}.

\paragraph{Quadruple approach.}

Here, the relationship between $(S_1,C_1)$ and $(S_2,C_2)$ is expressed as the relationship between two offsets in embedding space:
the offset from $S_1$ to $C_1$, and that from $S_2$ to $C_2$.
First, we compute the relationships between statutes and cases as vectors:

\begin{equation} \label{pairwise_1}
    g(S_1, \thinspace C_1) = \fsbert(S_1) - \fsbert(C_1)
\end{equation}

and likewise for $(S_2, C_2)$.
Then, we compute the cosine similarity between pairs of differences.
The score indicates whether these two sentences are analogical.
A higher similarity score indicates a higher degree of analogy.


\begin{equation} \label{quad}
   \mathrm{sim}((S_1,C_1), (S_2,C_2)) = \frac{ g(S_1,C_1) \cdot g(S_2,C_2) }{ || g(S_1,C_1) || \thickspace || g(S_2,C_2) ||} 
\end{equation}

\paragraph{Pair approach.}

As a comparison, illustrated in the top of Figure~\ref{fig:similarity_method}, we compute the cosine similarity of $(S_1, C_1)$ and $(S_2, C_2)$ directly.
For that, we embed the concatenation of statute and case using Sentence-BERT.
The similarity score is computed as:

\begin{center}
\begin{equation}\label{pair}
    \mathrm{sim}((S_1,C_1), (S_2,C_2)) = \frac{ \fsbert(S_{1}+C_{1}) \cdot \fsbert(S_{2}+C_{2}) }{ || \fsbert(S_{1}+C_{1}) || \thickspace || \fsbert(S_{2}+C_{2}) || }
\end{equation}
\end{center}

where ``$+$'' is string concatenation. As a final step, since our approach is based on computing similarity scores, we set a threshold to map similarity scores to binary labels.

\subsection{Binary Classification with T5-Large}
\label{sec:binary-classification}

Since we now have a dataset of legal analogy quadruples $(S_1, C_1, S_2, C_2)$ and their corresponding labels denoting whether it is analogical or not, we can run this task as a binary classification task.
For each classification example, we pass as input the string ${S_1+C_1+S_2+C_2}$. We expect as output a binary label, either $0$ (not analogy) or $1$ (analogy).
We fine-tune T5-Large on this binary classification task, using an MLP on top of the meanpool of the token representations.
In Section~\ref{sec:statutory-reasoning-as-retrieval-and-analogy}, we swap out T5-Large for bert-base-cased, to test the impact of the encoder.





\subsection{Statutory Reasoning as Retrieval and Analogy}

Connecting our analogy task to the original statutory reasoning task, we seek to answer the entailment problem by solving the analogy problem with the aid of retrieval mechanisms.
This is based on the idea that a new case should consult previous similar cases to make a judgment.
The following process is illustrated in Figure~\ref{fig:a2e}.

When given a test case, we first retrieve the most similar $k$ cases from the training set using a retrieval mechanism~--- either BM25 (Sparse Retrieval) or \ac{DPR}.
BM25 computes the concentration of the exact queried tokens in candidate texts, while \ac{DPR}~\cite{Karpukhin2020DensePR} represents both the query and texts using high dimensional vectors and computes a dot-product between them.
We concatenate the test case to each retrieved train example, so as to make $k$ instances of the analogy task, and then predict the analogy labels using our best-performing models for the analogy task.
Next, according to the entailment labels of train data examples and the analogy labels, we obtain $k$ predicted labels for the test example:
if the label is \emph{analogy}, the test example and the train example would have the same label; if no, then otherwise.
Finally, a majority vote based on the $k$ predicted entailment labels produces the final predicted entailment label on the test example.

\begin{figure}
\centering
    \includegraphics[width=0.65\textwidth]{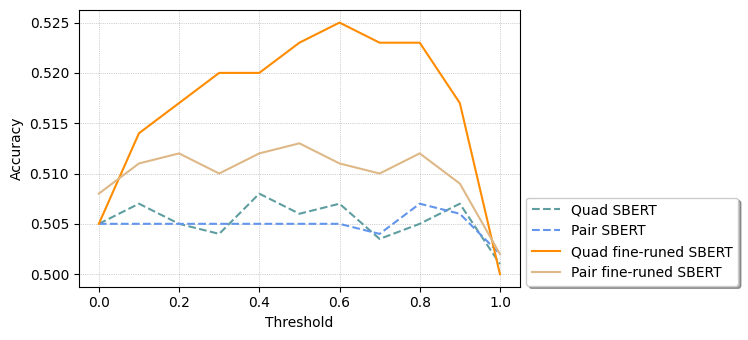}
    \caption{
    Prediction accuracy as a function of the similarity threshold, for different methods of computing similarity scores.
    A similarity score larger than the threshold will be labeled as \emph{analogy}, otherwise \emph{not analogy}.
    }
    \label{fig:accuracy}
\end{figure}


\section{Experiments and Results}

\subsection{Analysis with GPT models}
We conducted experiments on three models associated with GPT-3.5 and GPT-4, specifically focusing on different prompts, including zero-shot, few-shot, and chain-of-thought described in Section \ref{sec:gpt-3.5}. For zero-shot and few-shot settings, we additionally introduced a ``zero chain-of-thought''~\cite{kojima2022large} approach, i.e., using the prompt ``Let's think step by step'', to assess the model's performance on the analogy task. 
\\

\begin{table}[h!]
\centering
\caption{Performance of GPT-3.5 and GPT-4 on the analogy task. 5 different test sets of 100 items each are selected at random. We report the standard deviation of the accuracy across the 5 sets, using ``$\pm$''. ``Zero'' CoT indicates we used the prompt ``Let's think step by step'' \cite{kojima2022large}.}
\begin{tabular}{lc|ccc}
\toprule
&\textbf{Zero CoT} & \textbf{text-davinci-003} & \textbf{gpt-3.5-turbo-0301} & \textbf{gpt-4-0613} \\
\midrule
   \multirow{2}*{\textbf{zero-shot}} & Y    & 53 ± 2 & 49 ± 3 & 51 ± 2 \\
    ~ & N & 59 ± 4 & 54 ± 3 & 68 ± 3 \\
\midrule
    \multirow{2}*{\textbf{6-shot}} & Y    & 50 ± 3 & 49 ± 2 & 56 ± 4 \\
    ~ & N & 53 ± 2 & 53 ± 2 & 60 ± 3 \\
\midrule
    \shortstack{\textbf{6-shot hand-crafted}\\ \textbf{Chain-of-Thought}} & N & 59 ± 3 & 53 ± 3 & 57 ± 3 \\
\bottomrule
\end{tabular}

\label{tab:gpt3.5}
\end{table}


Our results are summarized in Table~\ref{tab:gpt3.5}. When evaluating various models on the analogy task, the exclusion of the prompt ``Let’s think step by step'' often results in improved or at least comparable performance across all models. Notably, the \textit{gpt-4-0613} model achieves highest performance in the zero-shot scenario without this specific prompt, underscoring the potential of newer model iterations. The consistent standard deviations seen in our results point to reliable model performance.

However, GPT models' capabilities on the analogy task appear to be somewhat limited. The model, while adept at generating context-rich text, seems better suited for the entailment task, which directly measures the relationships between statements, as evidenced by previous work \cite{blair2023can}. The introduction of the ``Let's think step by step'' prompt was conceived to guide the model in a more analytical direction. Yet, it may inadvertently have restricted its natural response mechanism, potentially compromising accuracy. This observation aligns with more recent empirical findings~\cite{DBLP:conf/acl/Shaikh0HBY23}, even if it is seemingly at odds with other prior research~\cite{kojima2022large}. 

\subsection{Vector Offset Method for Analogy Task}
\label{sec:experiments-vector-offset}

The generated quadruples were divided into training data and test data, and we ran four experimental settings.
Using the quadruple approach and pair approach mentioned before, we tested them on SBERT directly and after fine-tuning.
As shown in Figure~\ref{fig:accuracy}, we can see that all approaches yield close to 50\% accuracy.
Since the labels are balanced, the accuracy that would be had with a majority baseline is also 50\%.
Fine-tuned models achieve slightly higher accuracies:
52.5\% for the quadruple approach, and 51.2\% for the pair approach. 


Comparing to our classification-based approach described in Section~\ref{sec:binary-classification}, we performed a grid search over hyperparameters with Optuna~\cite{akiba19optuna}.
The best model achieves a dev accuracy of 53.3\%, and a test accuracy of 50.7\%.
This is lower than our approaches based on vector offset. Hyperparameter search and values can be found in Appendix~\ref{app:hyperparameters}.

\subsection{Statutory Reasoning as Retrieval and Analogy}
\label{sec:statutory-reasoning-as-retrieval-and-analogy}

\begin{figure}[h]
    \centering
    \includegraphics[width=\textwidth]{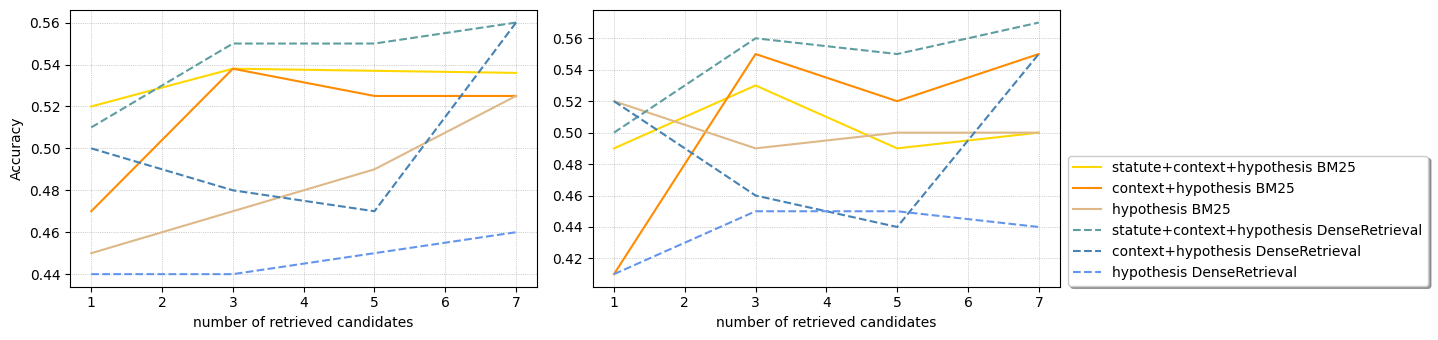}
    \caption{
    Statutory reasoning as retrieval and analogy with models from Section~\ref{sec:binary-classification}. Left: BERT-based model. Right: T5-Large-based model.
    }
    \label{fig:retrieval_result}
\end{figure}
\smallskip
We experiment with both BM25 and \ac{DPR}, and using our best-performing entailment models on the analogy task (described in Section~\ref{sec:binary-classification}).
As outlined in Section~\ref{sec:quadruple-generation}, each statutory example contains statute and case elements, with the latter further divided into context and hypothesis (Figure \ref{fig:a2e}).
We experiment with retrieving training examples based on different parts:
hypothesis, context+hypothesis or statute+context+hypothesis.
Additionally, we investigate the impact of varying $k$, representing the number of similar training data examples fetched for each test case. We opt for an odd number for simpler majority voting. 
Results are shown in Figure~\ref{fig:retrieval_result}.
 
For our T5-Large-based model described in Section~\ref{sec:binary-classification}, methods that make use of more information seem to achieve higher accuracies.
Both BM25 and \ac{DPR} methods using all of statute+context+hypothesis consistently have above-random accuracies.
In contrast, methods using context+hypothesis have less consistent accuracies, while still performing better on average than methods using hypothesis only.
Only \ac{DPR}-based methods that use statute+context+hypothesis consistently have accuracies better than random.
Results using bert-base-cased model are qualitatively consistent with those obtained with T5-Large.

The above experiments suggest that retrieving similar legal cases through \ac{DPR} followed by predictions made with analogy models could lead to better performance on statutory reasoning. Our best-performing configuration achieves an accuracy of 57\% in the statute+context+hypothesis setting. This is an improvement over textual-entailment-based models~\cite{Holzenberger2020ADF}, although not statistically significant, due to the small size of the test set.
Further investigation using other legal datasets would be helpful to determine the method's effectiveness.
\vspace{-10pt}
\section{Discussion and Conclusion}
\label{sec:discussion}
In this article, we reframed the SARA dataset as an analogy problem, generating 15,400 analogy samples from 176 entailment samples. Our exploration involved evaluating various methodologies, leveraging both \acp{LLM} and fine-tuning strategies tailored for BERT-like architectures. We observe the challenges and intricacies of the analogy task, a domain even powerful models like GPT find demanding. To gain insights into the results, we examined GPT outputs on 5 cases drawn at random and categorized them based on recurring patterns. See examples of typical behaviors of models in Appendix \ref{sec:model_behavior}.\\

\begin{table}[ht]
    \centering
    \caption{Summary of Model's Behavior on Analogical Reasoning Task. The result comes from both gpt-3.5-turbo-0301 (Turbo), text-davinci-003 (Davinci) and gpt-4-0613 (GPT-4).}
    \begin{tabularx}{\textwidth}{|c|X|c|c|c|c|}
        \hline
        \multirow{2}{*}{No.} & \multirow{2}{*}{Description}  & \multicolumn{4}{c|}{Occurrence} \\
        \cline{3-6}
        & & Davinci & Turbo & GPT-4 & \% \\
        \hline
        1 & Reasons whether S1 applies to C1 and S2 to C2 without genuine analogical reasoning. & 12 & 5 & 16 & 44.0\% \\
        \hline
        2 & Generates a straightforward (Yes/No) response or remains inconclusive. & 7 & 4 & 5 & 21.3\% \\
        \hline
        3 & Differentiates between ``definition'' type and ``specific situation'' documents. Often concludes negatively regarding document similarity. & 1 & 10 & 0 & 14.6\% \\
        \hline
        4 & Compares S1 to S2 and C1 to C2, occasionally demonstrating analogical reasoning as expected. & 2 & 2 & 3 & 9.3\% \\
        \hline
        5 & Other & 3 & 4 & 1 & 10.6\% \\
        \hline
    \end{tabularx}
    \label{table:GPT_behavior}
\end{table}



In Table \ref{table:GPT_behavior}, we have delineated discernible patterns in the model's responses. A primary observation is the model's inclination to determine if a Statute applies to a given Case, subsequently categorizing the outcomes based on this assessment. Beyond such rudimentary evaluations, our expectation, underscored by behavior No. 4, is for the model to engage in nuanced reasoning between Statutes and Cases, delving into their inherent similarities or dissimilarities. As highlighted in Section \ref{introduction}, there is an anticipation that the model will scrutinize both the definition and scope of statutes. 

Of the three models, both text-davinci-003 and gpt-4-0613 frequently simplify the analogy reasoning into two separate entailment problems for (S1, C1) and (S2, C2). In contrast, gpt-3.5-turbo-0301 is more inclined to delve into detailed comparisons between specific definitions and cases, thereby overlooking the reasoning of the underlying relationships.  Both scenarios suggest that LLMs may not adequately grasp the concept of an analogy relationship between two pairs. 

In our study, we observed that recasting tax law entailment within the framework of analogical reasoning presents challenges comparable to the original statutory reasoning task. Interestingly, models with suboptimal performance in analogy can still be adapted to serve the original statutory reasoning objective in our retrieval approach, yielding some improvement, albeit not reaching statistical significance.

Therefore, enhancing statutory reasoning could be achieved by integrating more sophisticated analogy models with advanced retrieval mechanisms, an approach that has the potential to surpass the results of previous methodologies. In future research, it may be beneficial to explore approaches to effectively combine these tools. For instance, when deciding which information is most useful, a weighted approach could be brought into the final voting, using the scores from BM25 and \ac{DPR}.

\vspace{-10pt}
\section*{Acknowledgements}

This work has been supported by the U.S. National Science Foundation under grant No. 2204926.
Any opinions, findings, and conclusions or recommendations expressed in this article are those of the authors and do not necessarily reflect the views of the National Science Foundation.
\vspace{-10pt}
\bibliographystyle{vancouver}
\bibliography{custom}

\begin{appendices}

\section{Quadruple Examples}
\label{sec:quadruple example}

\begin{table}[h!]
 \caption{An example of an analogical quadruple. }
\centering
\begin{tabular}{ |p{1cm}||p{10cm}|  }
 \hline
 Statute 1& Section 151. Allowance of deductions for personal exemptions (b) Taxpayer and spouse An exemption of the exemption amount for the taxpayer; and an additional exemption of the exemption amount for the spouse of the taxpayer if a joint return is not made by the taxpayer and his spouse, and if the spouse, for the calendar year in which the taxable year of the taxpayer begins, has no gross income and is not the dependent of another taxpayer.\\
 \hline
  Case 1&Alice and Bob have been married since 2 Feb 2015. Bob has no income for 2015. Alice can receive an exemption for Bob under section 151(b) for the year 2015.\\
  \hline
  Statute 2 &Section 63. Taxable income defined (b) Individuals who do not itemize their deductions In the case of an individual who does not elect to itemize his deductions for the taxable year, for purposes of this subtitle, the term ``taxable income'' means adjusted gross income, minus- (1) the standard deduction, and (2) the deduction for personal exemptions provided in section 151.\\
  \hline
  Case 2 &In 2017, Alice was paid \$33200. She is allowed a deduction under section 63(c)(1) of \$2000 for the year 2017, and no deduction under section 151. Alice takes the standard deduction. Under section 63(b), Alice's taxable income in 2017 is equal to \$31400.\\
 \hline
 Label & 0\\
 \hline
\end{tabular}

 \label{tab:example}
\end{table}

\begin{table}[h!]
 \caption{Example of Table~\ref{tab:example} further subdivided into context and hypothesis.}
\centering
\begin{tabular}{ |p{1cm}||p{5cm}|p{5cm}|  }
\hline
  Case & Context & Hypothesis\\
 \hline
  Case 1&Alice and Bob have been married since 2 Feb 2015. Bob has no income for 2015.& Alice can receive an exemption for Bob under section 151(b) for the year 2015.\\
  \hline
  Case 2 &In 2017, Alice was paid \$33200. She is allowed a deduction under section 63(c)(1) of \$2000 for the year 2017, and no deduction under section 151. Alice takes the standard deduction. & Under section 63(b), Alice's taxable income in 2017 is equal to \$31400.\\
 \hline
\end{tabular}

 \label{tab:case_example}
\end{table}

Our quadruple format is shown in Table~\ref{tab:example}. It is worth noting that for any case, we can further subdivide it into context and hypothesis as in Table~\ref{tab:case_example}.

\section{GPT-3.5 Prompts}
\label{sec:gpt3.5_prompts}

\begin{figure}[h!]
    \centering
    \includegraphics[width=\textwidth]{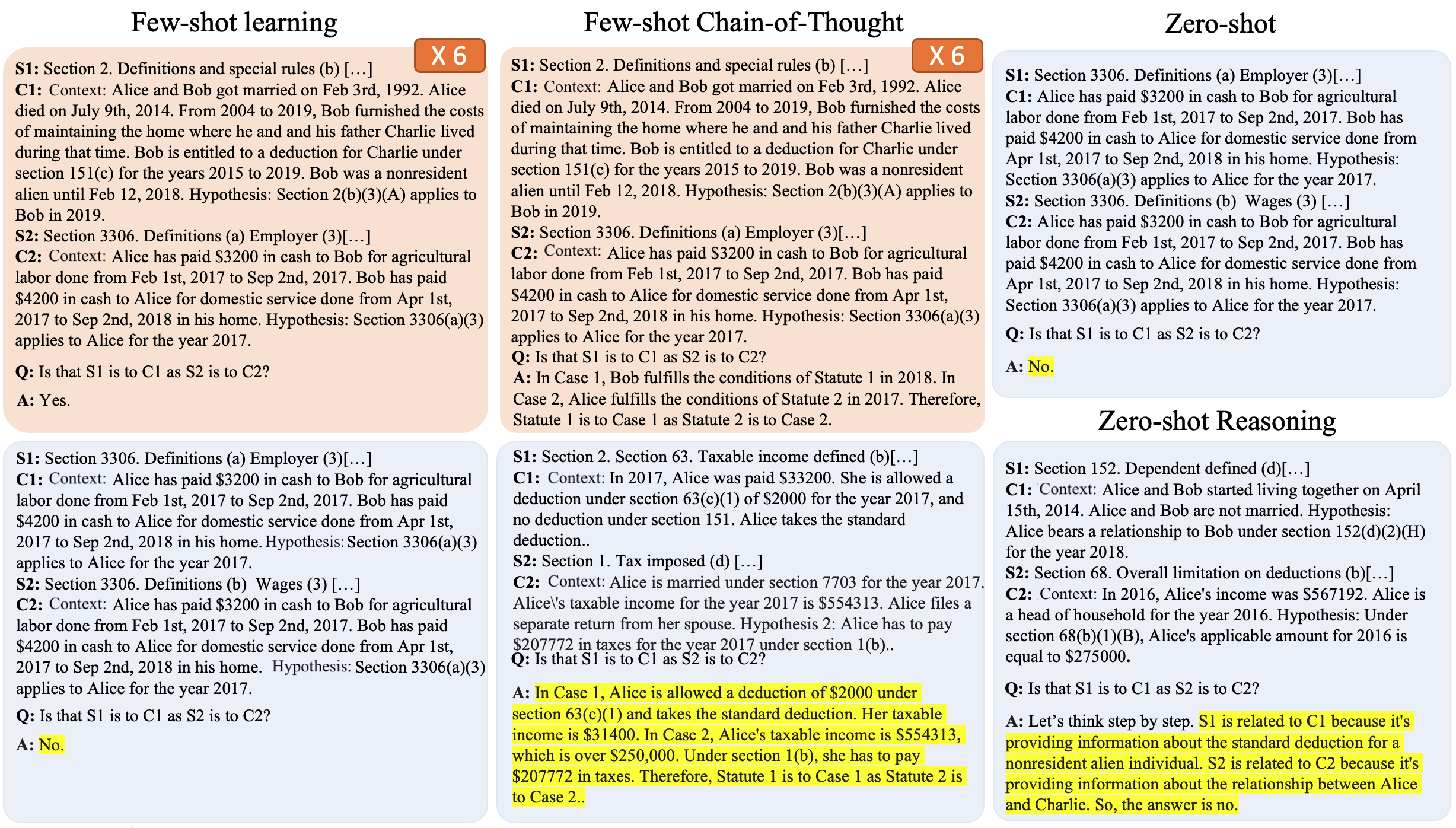}
    \caption{
    Prompts used in Section~\ref{sec:gpt-3.5} to process the analogy task with GPT models.
    The top orange boxes contain example quadruples. In the zero-shot setting there are no examples.
    The blue box at the bottom contains the test quadruple.
    Text highlighted in yellow is generated by a GPT model.}
    \label{fig:gpt3.5}
\end{figure}

Figure~\ref{fig:gpt3.5} shows examples of the prompts we used to produce the results in Table~\ref{tab:gpt3.5}.

\section{Hyperparameters}
\label{app:hyperparameters}

This section describes the hyperparameter range explored with Optuna in Section~\ref{sec:experiments-vector-offset}.
Table~\ref{tab:hyperparameter-search} shows the range explored for each hyperparameter, and Table~\ref{tab:hyperparameters-found} shows the hyperparameters of the best-performing model.

\begin{table*}[h!]
\scriptsize
\setlength{\tabcolsep}{.015\textwidth}
\centering
\caption{Hyperparameters in the hyperparameter search for T5-Large-based models.}
\begin{tabular}{p{.20\textwidth}p{.32\textwidth}p{.39\textwidth}}
\toprule
\textbf{Part of model}                     & \textbf{Hyperparameter} & \textbf{Value range} \\
\midrule
Encoder model & Number of trainable layers (counted from the last layer) & {0, 1, 2, 3, 4} \\
\midrule
MLP                                 & Number of units per layer         & \{128, 256, 512, 1024, 2048\} \\
                                  & Number of hidden layers        & \{0, 1, 2, 3, 4\} \\
\midrule
Trainer                           & Batch size              & \{8, 16, 32, 64, 128, 256\} \\
                                  & Learning rate           & [1e-7, 1e-3] \\
                                  & Learning rate warmup & [0, 0.5] \\
                                  & Learning rate scheduler & \texttt{constant\_schedule\_with\_warmup} \\
                                  & Optimizer               & \texttt{huggingface\_adamw} \\
                                  & Number of epochs to train for & 100 \\
                                  & Gradient clipping   & 1.0 \\
\bottomrule
\end{tabular}
\label{tab:hyperparameter-search}
\end{table*}

\begin{table}
\scriptsize
\setlength{\tabcolsep}{.015\textwidth}
\centering
\caption{Hyperparameters of the best T5-Large-based model.}
\begin{tabular}{p{.20\textwidth}p{.32\textwidth}p{.39\textwidth}}
\toprule
\textbf{Part of model}                     & \textbf{Hyperparameter} & \textbf{Value range} \\
\midrule
Encoder model & Number of trainable layers (counted from the last layer) & 4 \\
\midrule
MLP                                 & Number of units per layer         & - \\
                                  & Number of hidden layers        & 0 \\
\midrule
Trainer                           & Batch size              & 16 \\
                                  & Learning rate           & 5.9493e-06 \\
                                  & Learning rate warmup & 4.9602e-3 \\
\bottomrule
\end{tabular}

\label{tab:hyperparameters-found}
\end{table}

\section{Models' Behavior}
\label{sec:model_behavior}
Table \ref{tab:behavior_example} shows examples of typical behavior, for the same input analogical quadruple. The description of the behaviors can be found in Section \ref{sec:discussion}. The input to the model is shown at the bottom of the table.

\begin{table}[h!]
\centering
\caption{Examples of four typical outputs from the model.}
\scriptsize
\begin{tabularx}{\linewidth}{lX}
\toprule
\textbf{No.} & \textbf{Example}  \\
\midrule
1 & Statute 1 defines "wages" for the purpose of determining taxable income and includes remuneration paid in any medium other than cash, except for certain types of payments. Case 1 involves Alice paying Bob for work done and making payments into a retirement fund and life insurance. The hypothesis suggests that section 3306(b)(2)(C) applies to the life insurance payment. Based on these facts, it is reasonable to conclude that the hypothesis is supported by Statute 1. Statute 2 defines "employer" for the purpose of determining who is responsible for paying wages. It includes specific criteria for domestic service in a private home, local college club, or local chapter of a college fraternity or sorority. Case 2 involves Alice and Bob exchanging cash for agricultural labor and domestic service. The hypothesis suggests that section 3306(a)(3) applies to Bob for the year 2018. Based on these facts, it is reasonable to conclude that the hypothesis is supported by Statute 2. Therefore, the answer is yes, Statute 1 is to Case 1 as Statute 2 is to Case 2.
 \\
2 & Yes, Statute 1 is to Case 1 as Statute 2 is to Case 2.  \\
3 & No, Statute 1 does not directly apply to Case 1 as Statute 2 applies to Case 2. Statute 1 in Case 1 pertains to the definition of "wages" for the purposes of the chapter, specifically excluding certain payments made to employees or their dependents under a plan or system established by an employer on account of death. On the other hand, Statute 2 in Case 2 defines the term "employer" for domestic service in a private.\\
4 & Yes, Statute 1 to Case 1 is as Statute 2 is to Case 2. Both statutes pertain to specific definitions related to employment and wages, and the cases provide specific examples of employment situations to which the statutes could be applied. In both cases, the application of the statute hinges on the nature of the remuneration paid by the employer to the employee and the conditions under which it was paid.
  \\
\midrule
\textbf{Input} & Statute 1: Section 3306. Definitions (b) Wages For purposes of this chapter, the term "wages" means all remuneration for employment, including the cash value of all remuneration (including benefits) paid in any medium other than cash; except that such term shall not include- (2) the amount of any payment (including any amount paid by an employer for insurance or annuities, or into a fund, to provide for any such payment) made to, or on behalf of, an employee or any of his dependents under a plan or system established by an employer which makes provision for his employees generally (or for his employees generally and their dependents) or for a class or classes of his employees (or for a class or classes of his employees and their dependents), on account of- (C) death;Case 1:  Premise: Alice has paid \$45252 to Bob for work done in the year 2017. In 2017, Alice has also paid \$9832 into a retirement fund for Bob, and \$5322 into life insurance for Bob. Hypothesis 1: Section 3306(b)(2)(C) applies to the payment Alice made to the life insurance fund for the year 2017.Satute 2: Section 3306. Definitions (a) Employer (3) Domestic service In the case of domestic service in a private home, local college club, or local chapter of a college fraternity or sorority, the term "employer" means, with respect to any calendar year, any person who during the calendar year or the preceding calendar year paid wages in cash of \$1,000 or more for such service.Case 2:  Premise: Alice has paid \$3200 in cash to Bob for agricultural labor done from Feb 1st, 2017 to Sep 2nd, 2017. Bob has paid \$4200 in cash to Alice for domestic service in his home, done from Apr 1st, 2017 to Sep 2nd, 2018. Hypothesis 2: Section 3306(a)(3) applies to Bob for the year 2018.Question: Is Statute 1 to Case 1 as Statute 2 is to Case 2?\\
\bottomrule
\end{tabularx}
\label{tab:behavior_example}
\end{table}

\end{appendices}
\end{document}